%% file: main.tex
\documentclass[acmtog]{acmart}
\AtBeginDocument{%
  }

\copyrightyear{2024}
\acmYear{2024}
\setcopyright{rightsretained}
\acmConference[SA Conference Papers '24]{SIGGRAPH Asia 2024 Conference Papers}{December 3--6, 2024}{Tokyo, Japan}
\acmBooktitle{SIGGRAPH Asia 2024 Conference Papers (SA Conference Papers '24), December 3--6, 2024, Tokyo, Japan}
\acmDOI{10.1145/3680528.3687632}
\acmISBN{979-8-4007-1131-2/24/12}

\usepackage{graphicx}
\usepackage{bbding}

\usepackage{booktabs}
\usepackage{ulem}

\usepackage{multirow}
\usepackage{amsmath,bm}
\usepackage{url}
\usepackage[ruled]{algorithm2e} 

\SetAlFnt{\small}
\SetAlCapFnt{\small}
\SetAlCapNameFnt{\small}
\SetAlCapHSkip{0pt}
\usepackage{overpic} 
\usepackage{subcaption}
\usepackage{array}
\usepackage{float}

\usepackage{xcolor} 
\usepackage{colortbl}

\definecolor{shadecolor}{rgb}{0.92, 0.92, 0.92}
\definecolor{gtgray}{gray}{0.97}
\definecolor{mygray}{gray}{.88}

\definecolor{gray1}{gray}{.90}
\definecolor{gray2}{gray}{.92}
\definecolor{gray3}{gray}{.94}

\makeatletter
\def\hlinew#1{%
  \noalign{\ifnum0=`}\fi\hrule \@height #1 \futurelet
   \reserved@a\@xhline}
\makeatother
\usepackage[capitalize]{cleveref}
\crefname{section}{Sec.}{Secs.}
\Crefname{section}{Section}{Sections}
\Crefname{table}{Table}{Tables}
\crefname{table}{Tab.}{Tabs.}

\usepackage{array}
\newcolumntype{L}[1]{>{\raggedright\let\newline\\\arraybackslash\hspace{0pt}}m{#1}}
\newcolumntype{C}[1]{>{\centering\let\newline\\\arraybackslash\hspace{0pt}}m{#1}}
\newcolumntype{R}[1]{>{\raggedleft\let\newline\\\arraybackslash\hspace{0pt}}m{#1}}

\long\def\ignorethis#1{}
\definecolor{crimson}{rgb}{0.86, 0.08, 0.24}
\definecolor{green}{rgb}{0, 0.5, 0.25}
\definecolor{purple}{rgb}{0.75, 0, 1}
\definecolor{orange}{rgb}{1, 0.5, 0.25}
\definecolor{yellow}{rgb}{1, 1, 0}
\definecolor{new_blue}{rgb}{0, 0.5, 1}
\definecolor{new_cyan}{rgb}{0.10, 0.62, 0.57}

\begin{document}

\title{TextToon: Real-Time Text Toonify Head Avatar from Single Video}

\author{Luchuan Song}
\email{lsong11@cs.rochester.edu}
\affiliation{%
  \institution{University of Rochester}
  \country{USA}
}
\author{Lele Chen}
\affiliation{%
  \institution{University of Rochester}
  \country{USA}
}

\author{Celong Liu}
\affiliation{%
  \institution{Bytedance}
  \country{USA}
}

\author{Pinxi Liu}
\affiliation{%
  \institution{University of Rochester}
  \country{USA}
}

\author{Chenliang Xu}
\affiliation{%
  \institution{University of Rochester}
  \country{USA}
}

\begin{abstract}
We propose TextToon, a method to generate a drivable toonified avatar. Given a short monocular video sequence and a written instruction about the avatar style, our model can generate a high-fidelity toonified avatar that can be driven in real-time by another video with arbitrary identities. Existing related works heavily rely on multi-view modeling to recover geometry via texture embeddings, presented in a static manner, leading to control limitations. The multi-view video input also makes it difficult to deploy these models in real-world applications. To address these issues, we adopt a conditional embedding Tri-plane to learn realistic and stylized facial representations in a Gaussian deformation field. Additionally, we expand the stylization capabilities of 3D Gaussian Splatting by introducing an adaptive pixel-translation neural network and leveraging patch-aware contrastive learning to achieve high-quality images. To push our work into consumer applications, we develop a real-time system that can operate at 48 FPS on a GPU machine and 15-18 FPS on a mobile machine. Extensive experiments demonstrate the efficacy of our approach in generating textual avatars over existing methods in terms of quality and real-time animation. Please refer to our project page for more details: \url{https://songluchuan.github.io/TextToon/}.
\end{abstract}

%

\begin{teaserfigure}
\includegraphics[width=1.\textwidth]{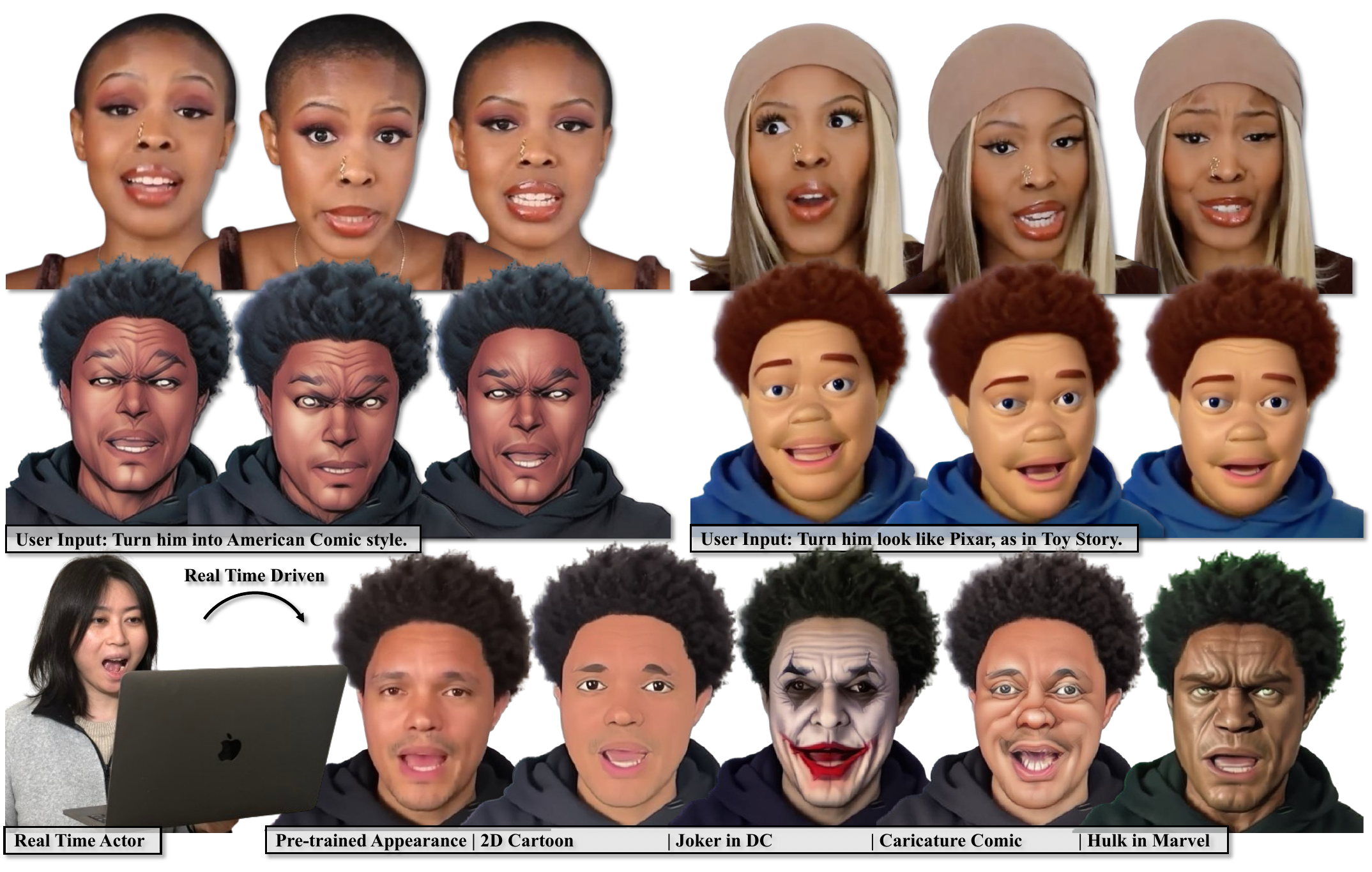}
\begin{small}
\end{small}
\caption{Our method takes a short monocular video as input (top) and animates the toonified appearance with synchronized expressions and movements using human-friendly text descriptions, e.g., "Turn him into an American Comic style" (middle). Moreover, our system achieves real-time animation (bottom), operating at 25 FPS (generation inference is about 48 FPS) on an NVIDIA RTX 4090 machine and 15 FPS on an Apple MacBook (M1 chip). All toonified faces (middle and bottom) are generated from the same pre-trained appearance model. Top block: Natural face$\copyright$\textit{Tee Noir} (CC BY). Middle and bottom: Natural face$\copyright$\textit{Trevor Noah} (CC BY).}
\label{fig:teaser}
\end{teaserfigure}

\maketitle

\input{629}

\end{document}

%% file: 629.tex
\section{Introduction}\label{sec:intro}

Generating high-quality toonified avatar videos is a long-studied topic in computer vision and computer graphics, with applications in daily life, social media, movies, and video game character creation. Recently, generative models have shown great promise in creating stylized portrait videos. For instance, VToonify~\cite{yang2022vtoonify} uses a learnable StyleGAN~\cite{karras2019style,karras2020training} to translate the specific portrait videos into different style via pre-collection images. AvatarStudio~\cite{pan2023avatarstudio} edits dynamic facial appearances by score distillation sampling~\cite{shi2023mvdream} from multi-view camera views. However, existing methods face challenges such as re-animation issues, low quality, and inference inefficiency, limiting their use in consumer applications.

An ideal toonification avatar generation system should meet the following principles: 
\begin{itemize}
\setlength{\itemsep}{5pt}
\setlength{\parsep}{0pt}
\setlength{\parskip}{0pt}
\item \textbf{Text Toonification}: Text is widely used for providing instructions in everyday activities,  requiring minimal expert knowledge. Consequently, a user-friendly system can generate the corresponding appearance based on user-input instructions, without pre-collection data or annotations. 
\item \textbf{Generalizability to the Driven Signal}: The toonified avatar generated should have the capability to be controlled by arbitrary identites with varying head motions and expressions.
\item \textbf{Rapid Adaptation and Real-time Animation}: The model should swiftly adjust to the desired style following user-input instructions within a few minutes. Once adapted, real-time animation should be achievable on consumer devices. 
\end{itemize}

To achieve the above principles, we propose an effective method named \textbf{TextToon} for head avatar toonification. Inspired by Tri-plane based neural volumetric representation, we use normalized orthographic rendering as the conditional Tri-plane inputs for 3D Gaussian points properties. The 3D Gaussian points, which control dynamic head expressions in the canonical space, are influenced not only by the linear deformations from 3DMM coefficients but also by the learnable Tri-plane features. However, since 3DMM is a linear model with limited facial expression capacity, we utilize the Tri-plane neural network structure to adaptively learn the mapping between the condition rendering and the corresponding stylized appearance in canonical space. Specifically, the normalized orthographic rendering map and 3DMM mesh deformed by expression provide a coarse representation of Gaussian point clouds. The fine-grained topology structure results from further displacement of the coarse points' positions by the learned Tri-plane features. For head motion, the estimated rotation and translation matrix are multiplied by each Gaussian point's property to achieve full movement.

Although the conditional Tri-plane-based 3D Gaussian Splatting significantly improves the accuracy of stylized facial expression representation, the toonification approaches still suffers from poor visual quality. For example, the inconsistencies in the diffusion-based Text-to-Image (T2I) editor~\cite{muller2022instant,zhang2023adding,shen2024controllable,wu2024text,aneja2023clipface} result in unclear and over-smoothed rendering outcomes~\cite{shao2023control4d}. To address this issue, we employ a patch-aware contrastive learning manner in the fine-tuning phrase. Specifically, the corresponding patches of the edited image and the target image are mapped to language embeddings with CLIP model~\cite{radford2021learning}, and contrastive constraints are imposed on the embeddings, which avoids the blur issue caused by direct pixel supervision.

\begin{figure*}[t]
  \centering
  \includegraphics[width=1.\linewidth]{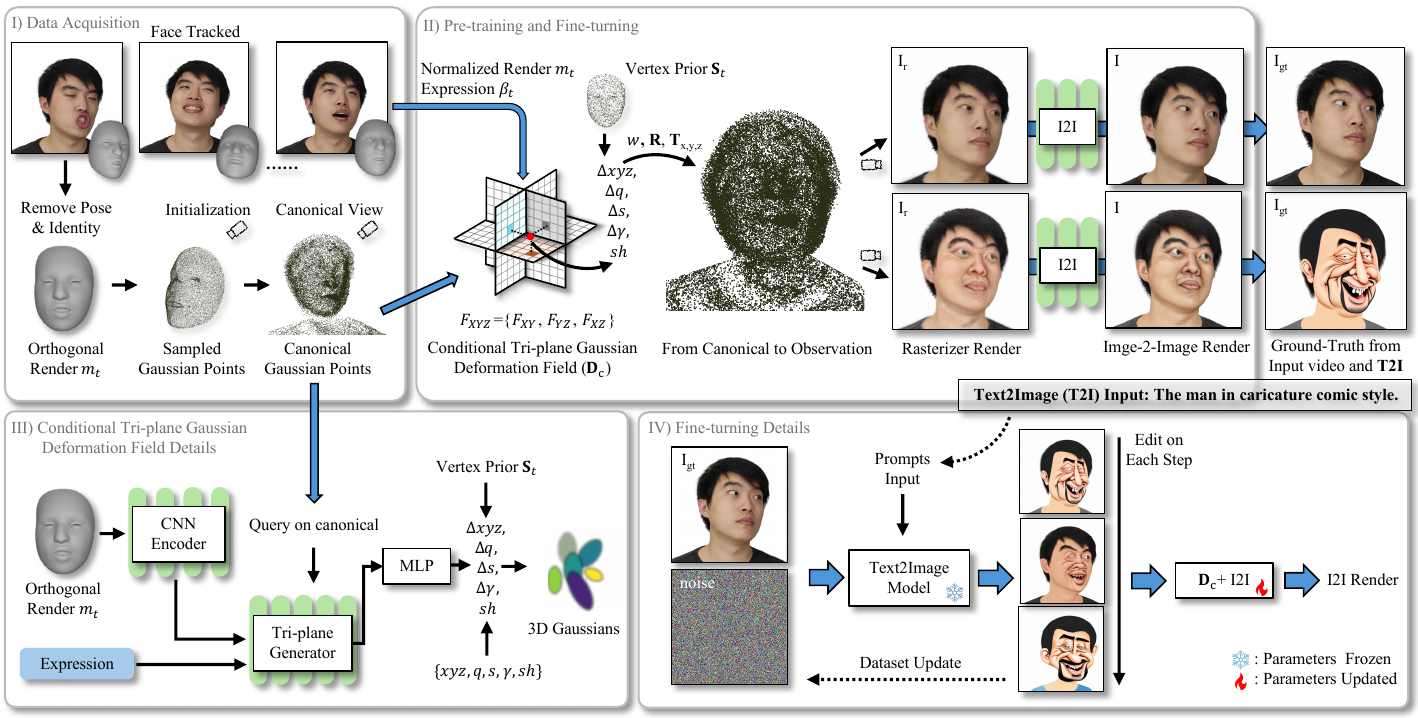}
  \caption{The overview of our methods. It takes a monocular video as input and tracks per frame, initializing the Gaussian point clouds using the tracked geometry from the first frame. We leverage the rigid transformation matrix ($\mathbf{R}, \mathbf{T}_{x,y,z}$) and a learnable lazy factor $w$ (in Sec.~\ref{Photo-Realistic Appearance Pre-training}) to transfer points from the canonical space to the observation space. The proposed conditional Tri-plane Gaussian Deformation Field $\mathbf{D}_c$ uses the normalized render map $m_t$, expression $\beta_t$ and vertex position $\mathbf{S}_t$ to predict the Gaussian properties deformation on each Gaussian points. Both the pre-training and fine-tuning phases share the same structure but target realistic appearance and T2I synthesized appearance, respectively. The details of conditional Tri-plane Gaussian Deformation Field and Text2Image editing are shown in III) and IV) respectively. Natural face$\copyright$\textit{Lizhen Wang et al.} (CC BY).}
  \label{fig_pipeline}
\end{figure*}

To accelerate the execution, we adopt a two-stage training strategy. At first, the framework is warmed-up on realism appearances by input single-view video. Then, we fine-tune the framework by the edited images from T2I module. The fine-tuning phrase usually takes five minutes. Benefit from the high inference speed of 3DGS, our pipeline could be easily accelerated to 48 FPS on single GPU machine and 15-18 FPS on mobile machine (\textit{e.g.} Apple MacBook M1 chip). Overall, our main contributions are as follows: 

\noindent \textbf{(1)} To the best of our knowledge, we present the first method for text-driven head avatar editing from monocular video. It leverages the state-of-the-art in 3D Gaussian Splatting and Text-to-Image (T2I) model to achieve high-quality editing of dynamic avatar.  

\noindent \textbf{(2)} We propose a conditional Tri-plane based Gaussian deformation field, which employs the canonical embedding to handle the nonlinear facial motion issue for toonification characters and significantly improve the animation accuracy.

\noindent \textbf{(3)} We design a real-time system capable of achieving inference speeds of over 48 FPS and pipeline about 25 FPS, with the 3DMM tracking algorithm being the efficiency bottleneck. Additionally, it is able to complete stylistic fine-tuning of special prompt in minutes. 

%

\section{Related Works}

\subsection{Parametrical Facial Model}

The parametric facial model~\cite{blanz2023morphable,bao2021high,yang2023asm,chai2022realy,FLAME:SiggraphAsia2017,brunton2014multilinear,neumann2013sparse,tewari2018self,tran2018nonlinear,wang2022faceverse,song2023emotional,grassal2022neural,hu2017avatar,song2021fsft,tran2019towards,zielonka2022towards} serves as an explicit facialization approach for reconstructing 3D face from RGB images. The facial shape \(S\) is articulated as:
\begin{equation}
S = S(\alpha, \beta) = \bar{S} + B_{id}\alpha + B_{exp}\beta,
\label{3DMM_can2world}
\end{equation}
where \(\bar{S} \in \mathbb{R}^{3E}\) denotes the average shape, with \(E\) representing the total number of vertices. \(B_{id}\) and \(B_{exp}\) are the PCA bases for identity and expression, respectively, and \(\alpha\), \(\beta\) are the corresponding parameters. The model facilitates spatial positioning adjustments through an Euler rotation matrix \(\mathbf{R} \in \mathbb{R}^{3\times3}\) and a translation vector \(\mathbf{T} \in \mathbb{R}^{3}\). The spatial coordinates \(S_{x,y,z}\) of the facial model is:
\begin{equation}
S_{x,y,z} = \mathbf{R} \cdot S(\alpha, \beta) + \mathbf{T}_{x,y,z},
\label{s_xyz}
\end{equation}
enabling the projection of the 3D facial model onto a 2D plane.

\subsection{Text-to-Image Generation}

The approaches for text-driven image (T2I) synthesis with diffusion models~\cite{ho2020denoising,dhariwal2021diffusion,radford2021learning}, they work on the static appearance editing in terms of content and style. Due to the powerful ability of diffusion model in style generation, the T2I is able to produce a wide range of visual with text prompts or other modalities. For example, Latent Diffusion Models (LDM)~\cite{rombach2022high} synthesis high-resolution images through text, controlnet~\cite{zhang2023adding} generates corresponding images under given control conditional maps and text prompts and InstructPix2Pix~\cite{muller2022instant} focuses on the edit the given images with prompts. 

\subsection{Text-to-Video Generation}

The Text-to-Video (T2V) is a further step toward T2I, it pay more attention to dynamic video generation rather that static images. Some state-of-the-art works~\cite{mei2023vidm,clark2024text,hong2022cogvideo,ho2022video} have achieved the continuous visual effect and aligned with the input prompts. These methods learn visual features corresponding to text embeddings from extremely large amounts of videos, and synthesize frames with optical flow as supervision.

\subsection{Video Transfer via StyleGAN}

Recently, the StyleGANs~\cite{karras2019style,karras2020training} show the outstanding performances in terms of video stylization. Some methods~\cite{richardson2021encoding,liu2022deepfacevideoediting} decode the aligned appearance of the corresponding style by editing the facial attributes latent codes. To achieve unaligned controllability, Stitch-Time~\cite{tzaban2022stitch} proposes an align-unalign-paste process to edit the in-the-wild videos. Furthermore, VToonify~\cite{yang2022vtoonify} and StyleGANEX~\cite{yang2023styleganex} incorporates the random geometric transformations for fine-tuning on the unaligned face images while improving the quality. 

\section{Method}

The pipeline of our proposed method is illustrated in Figure~\ref{fig_pipeline}. Given a single portrait video as input, we preprocess the video data using 3DMM estimation to generate normalized orthographic renderings $m_t$, expression parameters $\beta_t$, and corresponding vertex geometry $\mathbf{S}$ for each frame. Our method then applies the conditional Tri-plane Gaussian deformation field to edit the appearance in the canonical space and control expressions. The training phase is divided into two steps: (1) photo-realistic appearance pre-training and (2) text-driven appearance fine-tuning. Both steps share the same inputs but differ in objective functions. The (1) lies in the supervision of realistic appearance, and (2) is focused in the semi-supervised adaptation of style images.

\subsection{Data Acquisition}
\label{Data Acquisition}

We apply 3DMM tracking to generate the corresponding parameters (euler roatation $\mathbf{R}\in \mathbb{R}^{3\times3}$, translation vector $\mathbf{T}\in \mathbb{R}^{1\times3}$, facial identity $\alpha\in \mathbb{R}^{30}$ and expression $\beta\in \mathbb{R}^{52}$) for monocular inputs. The pose and identity removed geometry $\mathbf{S}\in \mathbb{R}^{1703\times3}$ is set with $\mathbf{R} = \mathbf{I}$, $\mathbf{T} = 1$ and $\alpha = 0$ in Equ.~\ref{s_xyz}.

For effectiveness, we directly solve for the analytical solutions of the 3DMM parameters with Gauss-Newton optimization on CPU, as in the face2face~\cite{thies2016face2face}. Specifically, we compute the Jacobian $\mathbf{J}$ related to the residual $\mathbf{r}$ of detected facial landmarks and projected landmarks, the preconditioned conjugate gradient (PCG) method is used for updating the parameters solutions $X$, as: 
\begin{equation}
\mathbf{J}^T \mathbf{J} \Delta P = -\mathbf{J}^T \mathbf{r},
\label{pcg}
\end{equation}
the $\Delta P$ is the updated scale of each step, we perform a four-step PCG in each Gauss-Newton optimization. In this way, we achieve the 3DMM estimation speed of $25\sim30$ FPS on a CPU, freeing up the GPU (if available) for backend processing.

\subsection{Tri-plane Gaussian Deformation Field}
\label{Conditional Tri-plane Gaussian Deformation Field}

The Tri-plane~\cite{chan2022efficient,song2024tri2planethinkingheadavatar,sun2022ide,shao2023control4d} has made significant improvement in 3D representation. It directly represents the density and color in NeRF through the stored three-plane features, which avoids the heavy rely on implicit features in connected MLP layers.

Inspired by the color and density decoded by Tri-plane, we propose the conditional Tri-plane Gaussian Deformation Field to decode the Gaussian properties. Considering that our setting is a person-specific neural rendering, the inputs are conditional embeddings instead of sampled noise. As shown in Figure~\ref{fig_pipeline}, it takes the normalized orthographic rendering maps $m_t$ and expression $\beta_t$ as input with vertex position $\mathbf{S}_t$ as prior to obtain the three plane features $\mathbf{F}_{XYZ} = \{\mathbf{F}_{XY}, \mathbf{F}_{XZ}, \mathbf{F}_{YZ}\}$. We build a Gaussian deformation decoder for the properties in 3DGS. Specifically, the points in Gaussian field can be represented by the offset on a position vector $xyz \in \mathbb{R}^{N \times 3}$, a quaternion vector $q\in \mathbb{R}^{N \times 4}$ and a scaling vector $s\in \mathbb{R}^{N \times 3}$ (please note the scale $s$ is distinguished from geometry $\mathbf{S}$), where the $N$ is the number of Gaussian points. Apart from that, each point has additional parameters: spherical harmonics (SH) $sh \in \mathbb{R}^{N \times 3}$ and opacity $\gamma \in \mathbb{R}^{N \times 1}$. The head canonical appearance Gaussian deformation field $\mathbf{D}_C$ can be formulated as:
\begin{equation}
\{xyz, q, s, sh, \gamma\} = \mathbf{D}_C(m_t, \beta_t, \mathbf{S}_t). 
\label{gdf}
\end{equation}

Considering that the expression motion of toonified face and the realism face are different, it is hard to directly fine-tune $\{xyz, q, s\}$ to achieve the toonified expression control. Therefore, we do not completely adopt the learned $xyz$ as the fine-grained face motion as previous methods~\cite{xu2023gaussianheadavatar,xiang2023flashavatar,qian2023gaussianavatars}. The vertex from coarse facial geometry is set as a prior Gaussian points for each frame. Specifically, we sample the vertices coordinate from each $\mathbf{S}_t$ and concatenate them with the learnable Gaussian points to jointly guide the rendering. %

\begin{figure}[t]
  \centering
  \includegraphics[width=1\linewidth]{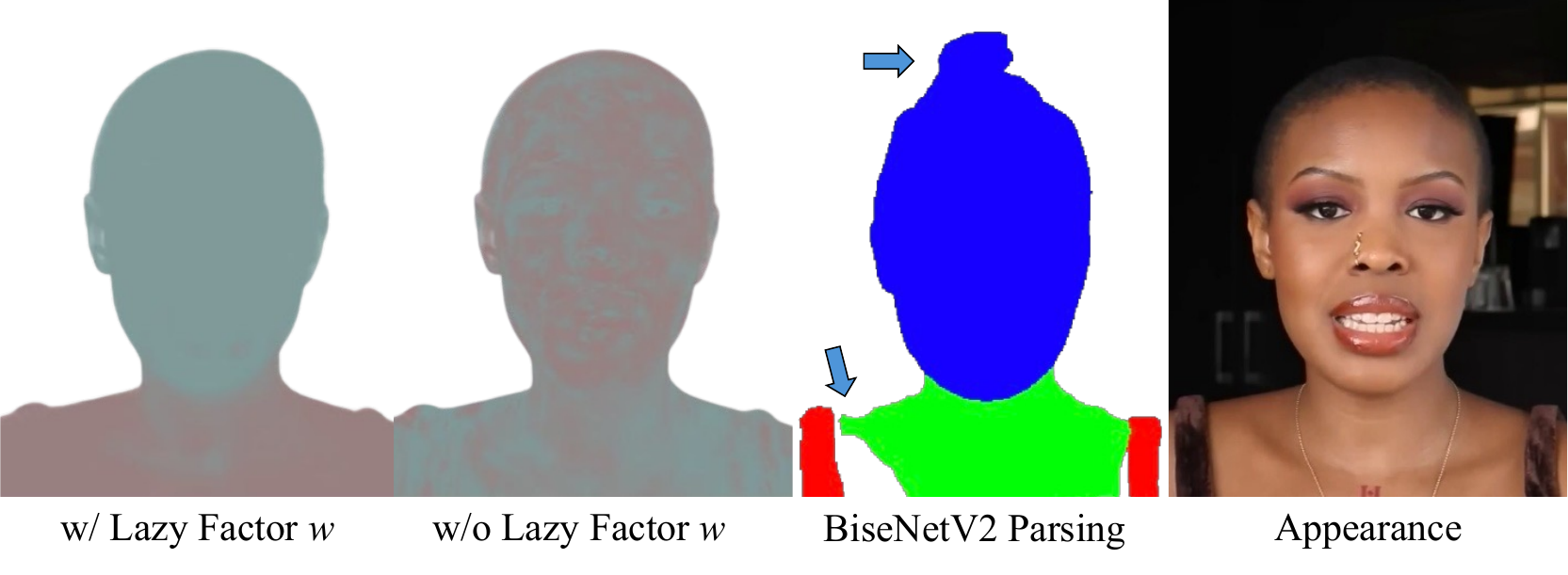}
  \caption{A visualization of the adaptively selected points via $w$. After introducing the lazy factor $w$, a soft boundary forms between the head and shoulders. Otherwise (w/o), they are mixed together and difficult to distinguish. It avoids mis-segmentation issues (indicated by the blue arrow). Natural face$\copyright$\textit{Tee Noir} (CC BY).}
  \Description{}
  \label{fig_rigid}
\end{figure}

\subsection{Photo-Realistic Appearance Pre-training}
\label{Photo-Realistic Appearance Pre-training}

\subsubsection{Non-rigid Motion Decoupling} In the 3DGS for dynamic scene, the rigid transformation matrix ($\mathbf{R}$ and $\mathbf{T}$) is applied to the point clouds from the canonical to the observation space. To avoid the artifacts from the non-rigid motion of the shoulder, SplattingAvatar~\cite{shao2024splattingavatar} and FlashAvatar~\cite{xiang2023flashavatar} remove the shoulder and only retain the face region (with neck). AD-NeRF~\cite{guo2021ad} leverages the pre-trained segmentation model to parse the face and train two NeRFs to represent different areas, which heavily relies on the accuracy of the segmentation model. To address this problem, our approach arises from two assumptions: I) The head and shoulder within the same structure, which cannot be modeled independently.  II) The shoulder is not completely fixed but exhibits the ``lazy" movement with much lower amplitude and frequency than head motion.

For assumption I), we initialize the shoulder from a cuboid structure point clouds, and optimize its properties in Gaussian splatting together with the face part. In assumption II), we introduce a learnable ``lazy" factor $w \in \mathbb{R}^{N\times 4}$ to control the shoulder movements. Specifically, the following rigid transformation is:
\begin{equation}
\begin{aligned}
xyz \cdot \mathbf{R} + \mathbf{T} & \rightarrow xyz', \\
\mathbf{R}_q \cdot q & \rightarrow q', \\
\end{aligned}
\label{rigid-1}
\end{equation}
the $\{xyz', q'\}$ is the position and quaternion vector in observation space, $\mathbf{R}^q$ is the quaternion corresponding to $\mathbf{R}$. Then, we apply the lazy vector $w$ for non-rigid adjustments to Eq.~\ref{rigid-1}:
\begin{equation}
\begin{aligned}
xyz \cdot \mathbf{R}\cdot w + \mathbf{T} & \rightarrow xyz', \\
\mathbf{R}_q \cdot q \cdot w & \rightarrow q'. \\
\end{aligned}
\label{rigid-2}
\end{equation}
The Eq.~\ref{rigid-2} can be divided into head and shoulder parts:
\begin{equation}
\begin{aligned}
xyz_1 \cdot \mathbf{R}\cdot w_1 + xyz_2 \cdot \mathbf{R}\cdot w_2 + \mathbf{T} & \rightarrow xyz'_1 + xyz'_2, \\
\mathbf{R}_q \cdot q_1 \cdot w_1 + \mathbf{R}_q \cdot q_2 \cdot w_2 & \rightarrow q'_1 + q'_2, \\
\end{aligned}
\label{rigid-3}
\end{equation}
$xyz_1, xyz'_1, q_1, q'_1, w_1$ are for face region and $xyz_2, xyz'_2, q_2, q'_2, w_2$ are for shoulder region, $w$ is the concatenation of $w_1$ and $w_2$. Then, the head moves rigidly and the shoulder moves lazily, so we have:
\begin{equation}
\begin{aligned}
w_1 \rightarrow \mathbf{I}_q, w_2 \rightarrow (\mathbf{R}^{-1})_q. \\
\end{aligned}
\label{rigid-4}
\end{equation}
The $w$ is learned according to the approximation in Eq.~\ref{rigid-4}, and $\mathbf{I}_q$ is the quaternion corresponding to the unit rotation matrix. %

As shown in Figure~\ref{fig_rigid}, the lazy factor $w$ is broadcast to each point in the 3D Gaussian to achieve non-rigid coupling of the head and shoulder parts with soft boundary. 

\subsubsection{Pre-training Objectives} We warm-up the Tri-plane Gaussian deformation field and image-to-image (I2I) module on the realism appearance. The objective function is divided into three parts, coarse loss, refine-grained loss, and regularization of $w$. 

The coarse loss $\mathcal{L}_{RGB}$ is euclidean distance supervision on pixel color of generated images ($I_r$ and $I$) and ground-truth ($I_{gt}$):
\begin{equation}
\begin{aligned}
\mathcal{L}_{RGB} = ||I_r - I_{gt}||_1 + ||I - I_{gt}||_1. \\
\end{aligned}
\label{lrgb}
\end{equation}

The refine-grained loss $\mathcal{L}_{LPIPS}$ is defined by the LPIPS~\cite{johnson2016perceptual} between output images ($I$) from I2I module and ground-truth:
\begin{equation}
\begin{aligned}
\mathcal{L}_{LPIPS} = \text{LPIPS}(I, I_{gt}). \\
\end{aligned}
\label{lvgg}
\end{equation}

At last, the regularization of $w$ is: 
\begin{equation}
\begin{aligned}
 ||w \cdot w^T_c - \mathbf{I}_q||_2, \\
\end{aligned}
\label{rw0}
\end{equation}
where $w_c \in \mathbb{R}^{N\times4}$ is a vector and equal to $[\mathbf{I}_q, \mathbf{R}_q]$, where $w = [w_1, w_2]$ is set by the face and shoulder point cloud. The total objective function is:
\begin{equation}
\begin{aligned}
\mathcal{L}_{RGB} + \mathcal{L}_{LPIPS} + \lambda \cdot ||w \cdot w^T_c - \mathbf{I}_q||_2, \\
\end{aligned}
\label{rws}
\end{equation}
the $\lambda$ is a hyperparameter and it is equal to $1e^{-3}$ in our work.  

\subsection{Text-Driven Appearance Fine-tuning}
\label{Text-Driven Appearance Fine-tuning}
Text-driven appearance is fine-tuned with the images edited by T2I model~\cite{brooks2023instructpix2pix} instead of the realistic appearance from video frames, as shown in Figure~\ref{fig_pipeline} IV). To achieve high-quality toonify editing and address inconsistency issues by T2I, the objective function focuses on feature distance rather than pixel distance.

Inspired by NeRF-Art~\cite{wang2023nerf}, we employ a patch-aware contrastive loss. Specifically, we randomly select three patches of size $224\times224$ from the edited image and encode each patch into the language feature space using the CLIP model~\cite{radford2021learning}. In each iteration, positive samples for contrastive learning are constructed from the edited image patches $\pi(\mathbf{P}_j)$, where $\pi$ is the pre-trained CLIP model for textural feature embedding and $\mathbf{P'}_j$ is a patch from the edited images. Negative samples are built from predefined negative prompts $\pi(\text{neg})$, where $\text{neg}$ is randomly selected from a set of negative prompts such as {\textit{\{“Animal, Low quality, Blurry, Low res, Long neck, ...etc"\}}. During optimization, we aim to shorten the distance between $\mathbf{P'}_j$ and the positive samples while pushing them away from the negative descriptions. The patch-aware contrastive learning objective function is: 
\begin{equation}
\begin{aligned}
    \mathcal{L}_{CON} = -\sum_{\mathbf{P}\in \text{I}} \text{log} [\frac{\text{exp}( \pi(\mathbf{P'}_j) \cdot \pi(\mathbf{P}_j))}{\text{exp}(\pi(\mathbf{P'}_j) \cdot \pi(\mathbf{P}_j)) + \text{exp}(\pi(\mathbf{P}_j) \cdot \pi(\text{neg}))}], 
\end{aligned}
\label{con}
\end{equation}
where the $\mathbf{P}_j$ is the corresponding rendered image patches from Gaussian splatting. From Eq.~\ref{con}, we avoid measuring the pixel distances and refer to text feature distances. 

Taking over the objective functions in Sec.~\ref{Photo-Realistic Appearance Pre-training}, the objective functions for fine-tuning are:
\begin{equation}
\begin{aligned}
\mathcal{L}_{LPIPS} + \lambda_1 \cdot \mathcal{L}_{CON} + \lambda_2 \cdot ||w \cdot w^T_c - \mathbf{I}_q||_2, \\
\end{aligned}
\label{rw4}
\end{equation}
where the $\lambda_1$ and $\lambda_2$ are set to $1e^{-3}$ in our work.

The overall objects remain similar to Eq.~\ref{rws}, the difference is that we apply $\mathcal{L}_{CON}$ instead of $\mathcal{L}_{RGB}$. The objective is to accentuate the high-frequency features of the toonified appearance, such as teeth and hair, while avoiding issues of over-smoothing.

\section{Experiments}

\subsection{Implementation Details}

We conduct experiments with $8$ subjects monocular videos from the public datasets PointAvatar~\cite{zheng2023pointavatar}, StyleAvatar~\cite{wang2023styleavatar}, and InstantAvatar~\cite{zielonka2023instant}. Apart from that, we include two videos from self-recordings and YouTube to evaluate performance in the wild. All videos are presented at $25$ FPS with a resolution of $512 \times 512$. We take an average length of $1000$-$6000$ frames for training (about 80\%) while the test dataset includes frames with novel expressions and poses (about 20\%). For each frame, we apply the RobustVideoMatting~\cite{lin2022robust} to remove the background. The number of Gaussian points is set to $1e^4$, and the number of prior vertex points is $1703$. During pre-training and fine-tuning phases, the learning rate for Gaussian properties $\{xyz, q, s, \gamma, sh\}$ are $\{2e^{-5}, 1e^{-3}, 5e^{-3}, 5e^{-2}, 2.5e^{-3}\}$. The learning rate for $w$, I2I module, Tri-plane generator and encoder are $1e^{-3}$, $1e^{-2}$, $1e^{-3}$ and $1e^{-3}$ respectively. We employ the InstructPix2Pix~\cite{brooks2023instructpix2pix} as T2I module with 20 denoise steps to edit the frames. 
It takes about 40k iterations for photo-realistic appearance pre-training, and $200-500$ iterations for text-driven appearance fine-tuning. 

We employ the MLPs with dimensions [32+14] for the decorder of Tri-Plane, these account for 11-dimensions for xyz (Gaussian points position), Rotation, Scale and Opacity, 3-dimensions for Colors, respectively. And other components are shown and discussed in Section~\ref{Discussion and Conclusion} and Figure~\ref{fig:network_structure}.

\begin{figure}[t]
  \centering
  \includegraphics[width=1.\linewidth]{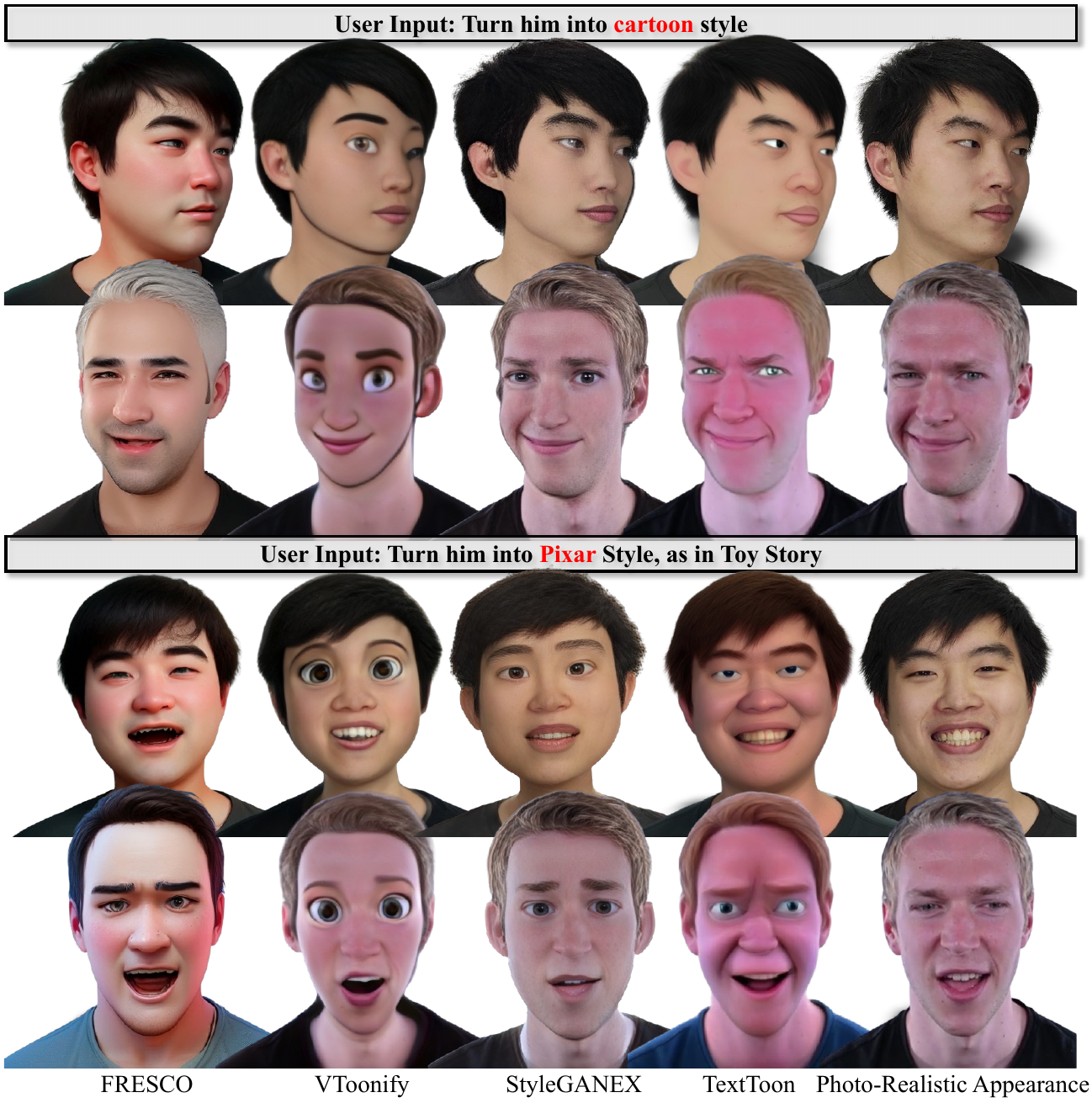}
  \caption{The perceptual evaluation of our method and baselines. For the sake of fairness and to avoid cherry-picked results, we adopt Pixar/cartoon stylization models provided by StyleGANEX and VToonify, and align them with our prompts. The style strength of VToonify is set to $0.7$. Natural face$\copyright$\textit{Lizhen Wang et al.} (CC BY), and $\copyright$\textit{Wojciech Zielonka et al.} (CC BY).}
  \label{fig:compare}
\end{figure}

\subsection{Baseline}
Given the challenges associated with text-driven avatar stylization, including video continuity, monocular-limited input, and synchronized re-animation movements, few methods exactly handle this task. Therefore, we select several related works for comparison, as,
\begin{itemize}
\setlength{\itemsep}{5pt}
\setlength{\parsep}{0pt}
\setlength{\parskip}{0pt}
\item \textbf{VToonify}~\cite{yang2022vtoonify}: It is a StyleGAN-based for pre- collection and annotation style images transfer to  unaligned faces. We take the in-the-wild portrait movement to generate the corresponding toonified images. 
\item \textbf{StyleGANEX}~\cite{yang2023styleganex}: The StyleGANEX is also based on unalignment StyleGAN, which brings more flicker suppression and improved video consistency than VToonify.
\item \textbf{FRESCO}~\cite{yang2024fresco}: The setting is similar to StyleGANEX and VToonify. But the difference is that it does not need to the pre-collection style images but adopts prompts and stable diffusion to edit the video style with optical flows.
\end{itemize}
We also acknowledge other related methods, such as AvatarStudio~\cite{pan2023avatarstudio}. However, due to its multi-camera setup (differing from the single view) and the lack of sufficient implementation details, we do not include it in the detailed comparison.

\subsection{Numerical Evaluations}

Due to the absence of ground-truth data, it is challenging to evaluate text-driven visual edits numerically. At here, we take two criteria for numerical evaluations, one is from quantitative measurement, the other is from human assessment. 

\subsubsection{Quantitative Metrics} It is based on three aspects. (1) Blind / Referenceless Image Spatial Quality Evaluator (BRISQUE): The BRISQUE~\cite{mittal2011blind} is a non-reference image quality assessment method, and is used to directly assess the quality of edited images. (2) Text-Image Consistency (CLIP-D)~\cite{haque2023instruct}: The CLIP-D is the embedding between text and image to calculate cosine similarity for each pair via CLIP. A higher CLIP-D score indicates a closer match between the generated appearance and the text. (3) Standard Deviation for video stylization (STD): We measure the CLIP features standard deviation of each frames across the video, which represents the stability of style in the video. (4) Frame Per Second without 3DMM Tracking (FPS w/o Track): We evaluate model inference speed on a single NVIDIA RTX4090 GPU. 

\subsubsection{User Study} We sample $10$ different identities each one with $8$ styles ($80$ videos for self-reenactment and 50 videos for cross-reenactment), and invite $32$ attendees from the Amazon Mechanical Turk (MTurk) to join the study. The Mean Opinion Scores (MOS) rating protocol is adopted for evaluation and the attendees are required to rate the generated videos from four aspects: (1) \textbf{IP} (Identity Preservation): Do you agree the identity is well-preserved in the stylized video? , (2) \textbf{TP} (Text Preservation): Do you agree the text description is well-presented in the stylized video?, (3) \textbf{MS} (Motion Synchronization): Do you agree the head motion in stylized video is synchronized with source one? and (4) \textbf{VQ} (Video Quality): Do you agree the overall video quality is good, e.g. frame quality, temporal consistency \textit{etc}? There is a $5$-point Likert scale for each term, range from $1$-$5$ corresponds to strongly disagree, disagree, don't know, agree and strongly agree respectively (the closer to $5$ the better). 

\begin{figure}[t]
  \centering
  \includegraphics[width=1.\linewidth]{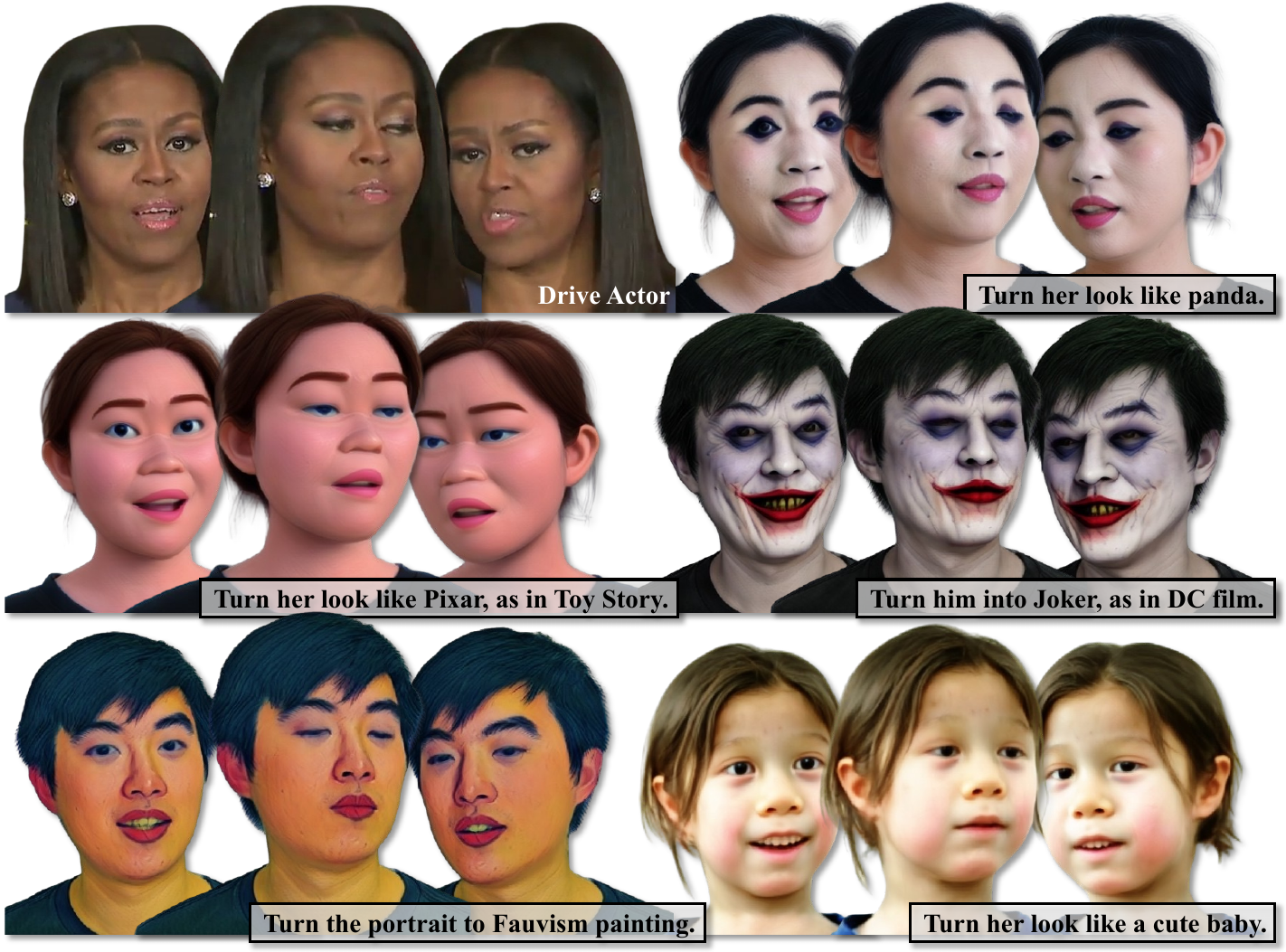}
  \caption{Visualization of cross-identity driven results. The drive actor is captured in the wild ($\copyright$Obama White House Daily), and the toonified avatar from a different identity is synchronized by facial expressions and poses.}
  \label{fig:cross}
\end{figure}

\begin{table}[t]
\footnotesize
\begin{center}
\setlength{\tabcolsep}{1.mm}
{
\begin{tabular}{ccccc|cccc}
\hlinew{.8pt}
\multirow{3}{*}{Methods} &BRISQUE$\downarrow$ &CLIP-D$\uparrow$ &STD$\downarrow$  &FPS$\uparrow$ &\textbf{IP}$\uparrow$ &\textbf{TP}$\uparrow$ &\textbf{MS}$\uparrow$ &\textbf{VQ}$\uparrow$\\
\cline{6-9}
&($\times 10^{-2}$) &($\rightarrow 1$) &($\rightarrow 0$) &(w/o Track) &\multicolumn{4}{c}{($\rightarrow 5$)}\\
\cline{2-9}
&\multicolumn{4}{c|}{Quantitative Results}&\multicolumn{4}{c}{User Study}\\
\hline
VToonify & 0.62 & 0.17  & 0.171  & \underline{14} & 3.1 & \textbf{4.5} & \underline{4.2} & \underline{3.9}  \\
StyleGANEX & \underline{0.54} & 0.10 & 0.195 & 8.4  &   \textbf{4.3} & 2.9 & 3.5 & 3.2    \\
FRESCO &0.58& \underline{0.21} & 0.227 &  0.5 &  2.9 & 3.5 & 3.0 & 3.8     \\
\rowcolor{mygray} \textbf{TextToon} &\textbf{0.49} &\textbf{0.25} &\textbf{0.130}  &\textbf{48} & \underline{3.8} & \underline{4.3} & \textbf{4.7} &\textbf{4.1}  \\
\hlinew{.8pt}
\end{tabular}}
\vspace{.3cm}
\caption{(1) Left: Quantitative evaluations with the baseline methods. The $\downarrow$/$\uparrow$ indicates lower/higher values of better performance. The best results are in bold, and the second-best are underlined. (2) Right: The 5-point Likert scale for user study (closer to 5 indicates better performance).}
\label{table_1}
\end{center}
\end{table}

\begin{figure}[t]
  \centering
  \includegraphics[width=1.\linewidth]{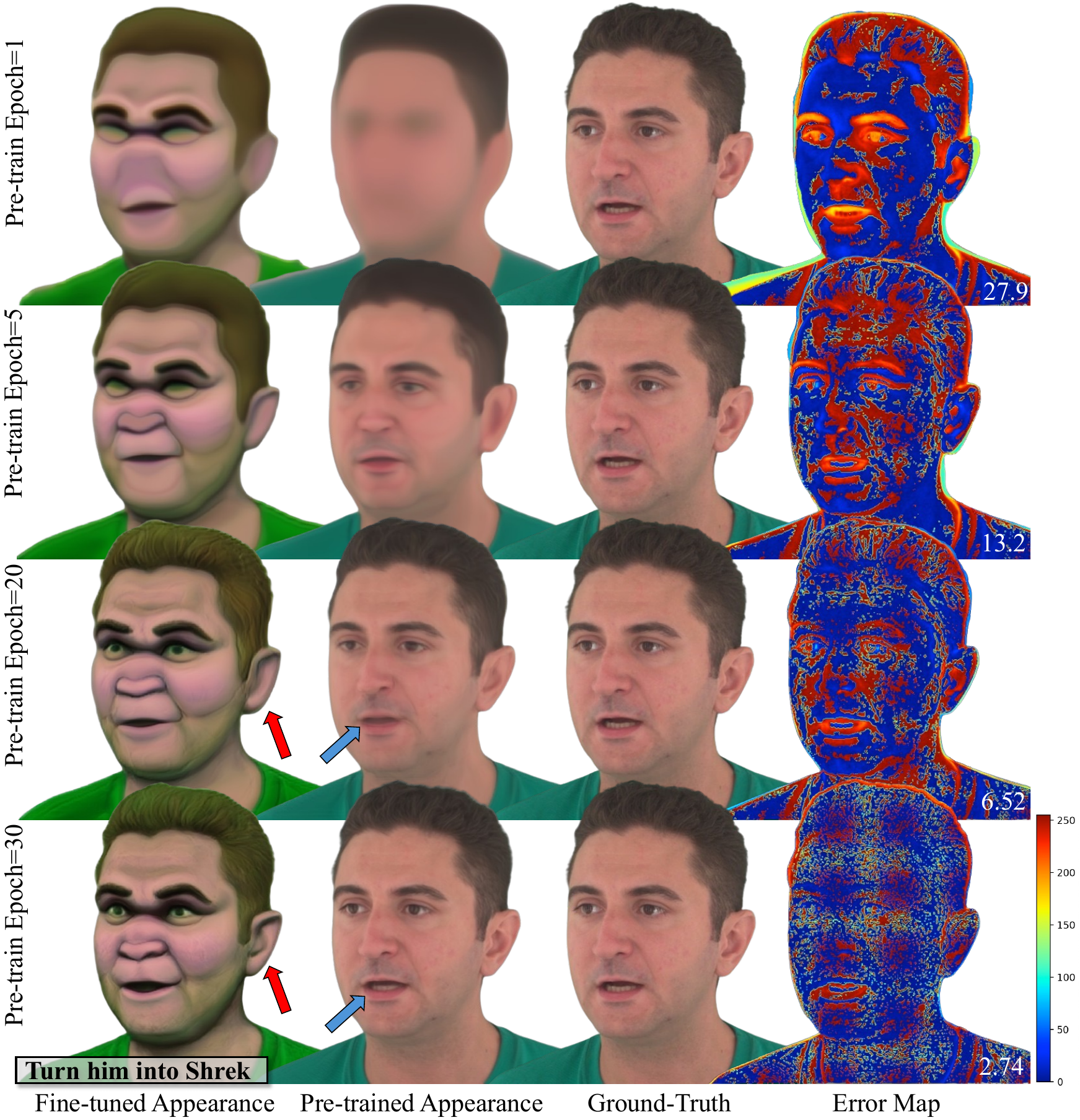}
  \caption{The relationship of pre-trained and fine-tuned appearance. We list the different correspondences, a better pre-trained model results in a more detailed fine-tuned appearance (the pre-training iterations are not constant, but the fine-tuning iterations are consistent). The error maps show pixel Euclidean distance in RGB (color in [0, 255]). A lower mean error (white number) indicates a better pre-trained appearance. Please refer to the arrows for facial details. Natural face$\copyright$\textit{Yao Feng et al.} (CC BY).}
  \label{fig:em}
\end{figure}

\subsection{Comparison with Baseline Methods}

The comparison results are shown in Table~\ref{table_1} and Figure~\ref{fig:compare} respectively. From Table~\ref{table_1}, our method achieves almost the best or second best results in user study and quantitative evaluations. The VToonify~\cite{yang2022vtoonify} is still a powerful baseline, which leads the way in terms of style preservation (\textbf{TP}) and FPS. As the meantime, our method achieves the most consistent style across one video from STD. 

The perception comparisons are shown in Figure~\ref{fig:compare}, our approach achieves a better performance than the excessive style of VToonify or the source identity preservation of StyleGANEX. It is worth noting that stylization assessment is based on personal perception awareness. We acknowledge the superiority of baseline methods in different aspects, but our method is also better than them in terms of \underline{user-friendly text input}, \underline{fast style fine-tuning}, \underline{inference speed} and \underline{re-animation}. Additionally, we present the results of re-animation are shown in Figure~\ref{fig:cross}, where the actor is in the top-left in the Figure~\ref{fig:cross} and the rest are driven text-toonification appearances.

\begin{figure}[t]
  \centering
  \includegraphics[width=1.\linewidth]{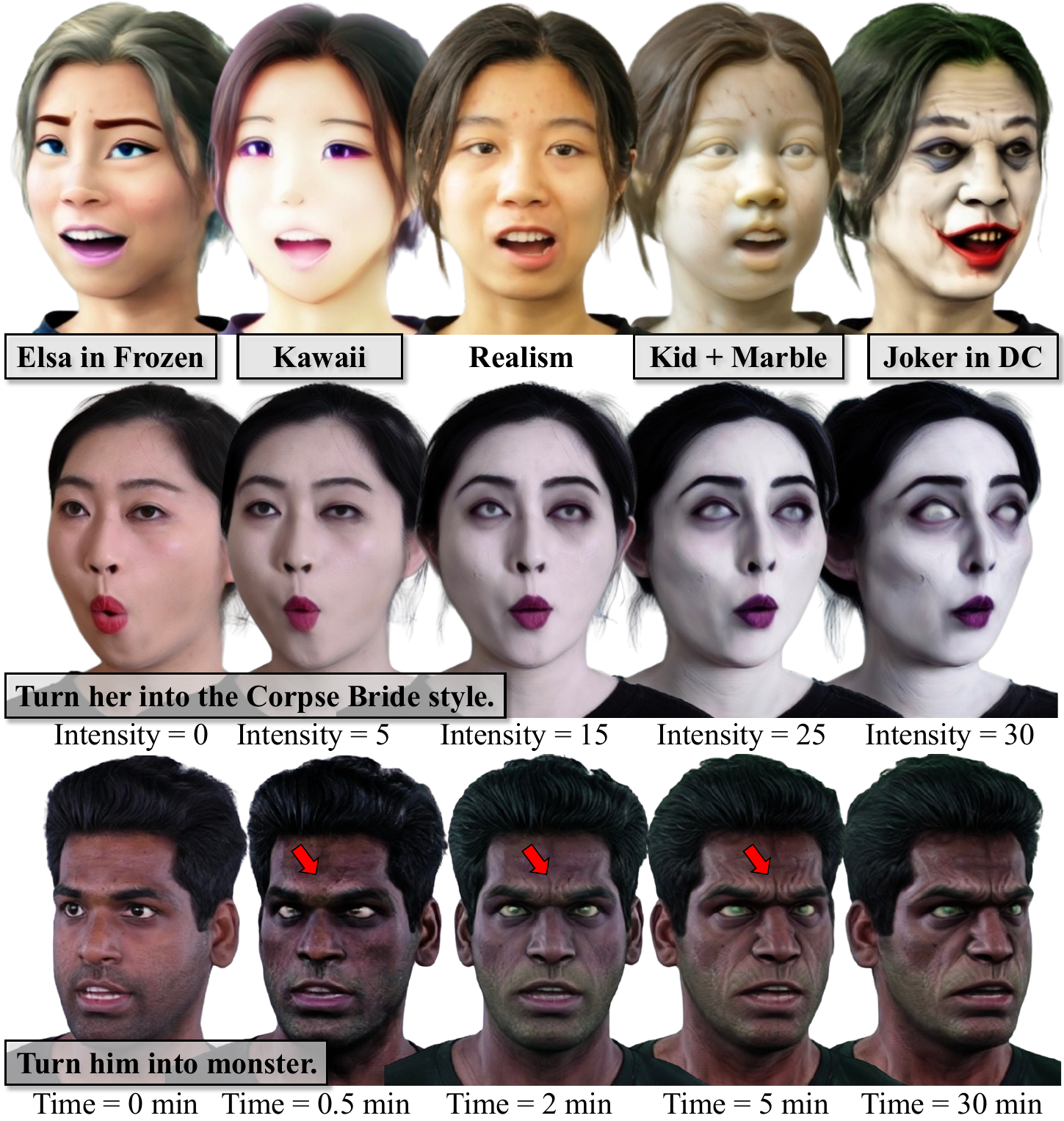}
  \caption{The visualization of three ablation studies (from top to bottom) results from \textbf{multi-view}, the multi-view results are only learned from monocular inputs. (1) Top: The fine-tuned appearance by our method with different prompts. (2) Middle: The visualization of stylization intensity, where intensity = 0 represents the source appearance. (3) Bottom: The style trajectory over fine-tuning time, with Time = 0 min representing the source appearance. Please refer to the arrow for details. At the meantime, we present the multi-view visualization results at here for perception assessment, the multi-view appearance are learned from the monocular inputs. Natural face$\copyright$\textit{Yufeng Zheng et al.} (CC BY), and $\copyright$\textit{Wojciech Zielonka et al.} (CC BY).}
  \label{fig:m-view}
\end{figure}

\begin{figure}[t]
  \centering
  \includegraphics[width=1.\linewidth]{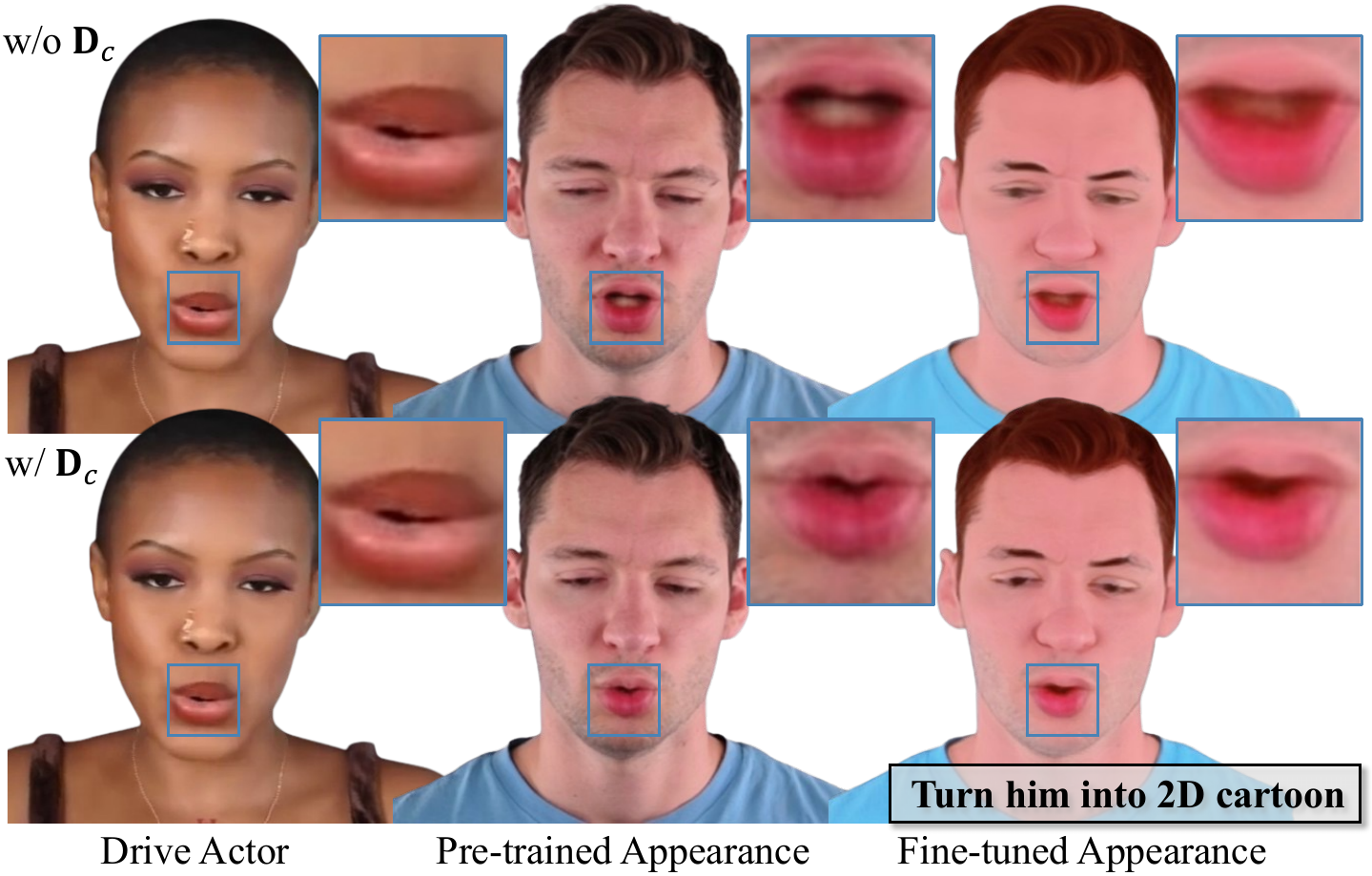}
  \caption{We visualize the conditional Tri-plane Gaussian deformation field $\mathbf{D}_c$ using a cross-identity driven approach. Utilizing $\mathbf{D}_c$ (w/) helps manage more complex mouth shapes compared to not using $\mathbf{D}_c$ (w/o). This improvement occurs throughout both the pre-training and fine-tuning phases. We highlight the mouth details for better evaluation. Natural face$\copyright$\textit{Wojciech Zielonka et al.} (CC BY).}
  \label{fig:D_c}
\end{figure}

\begin{figure}[t]
  \centering
  \includegraphics[width=1.\linewidth]{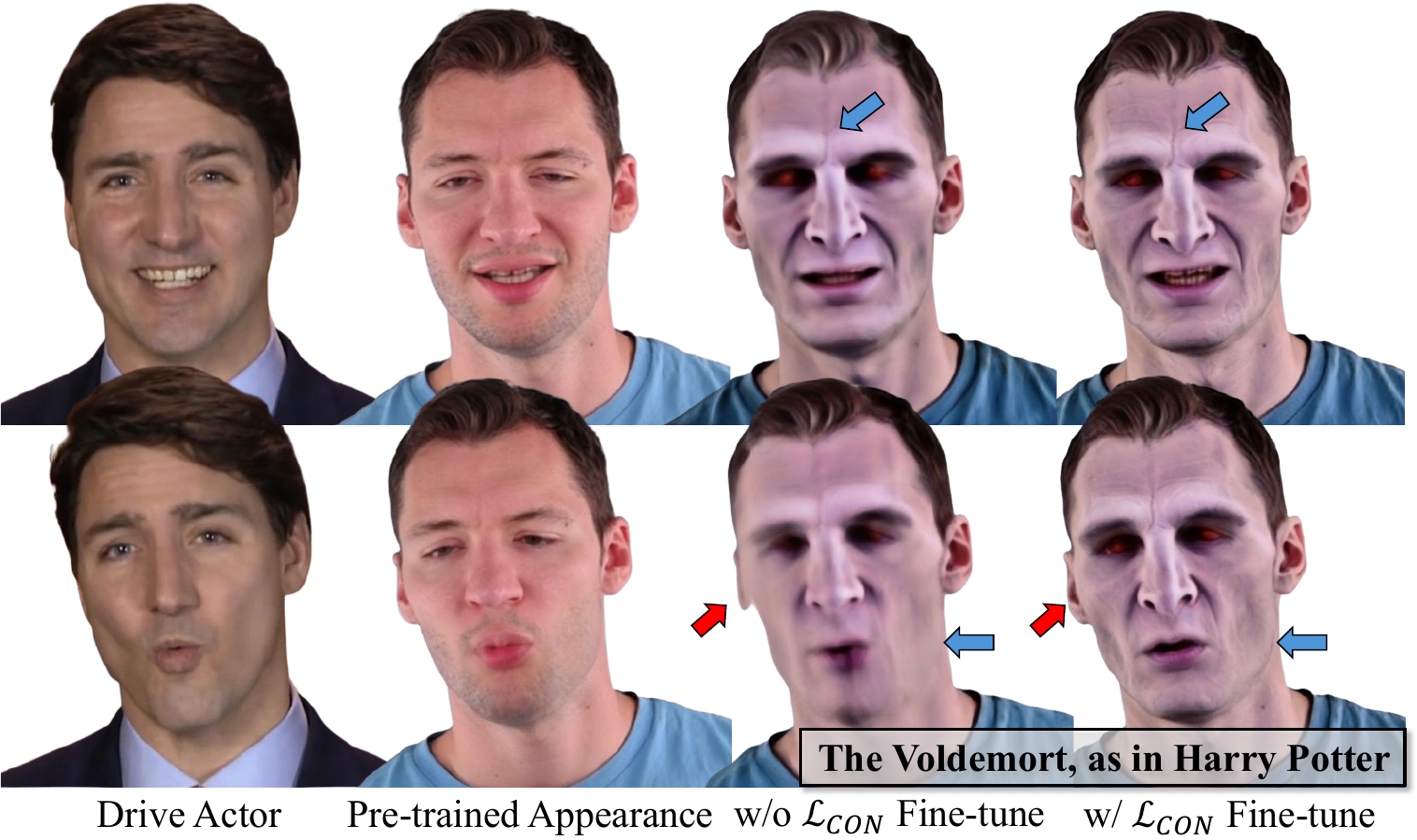}
  \caption{The visualization of the ablation study $\mathcal{L}_{CON}$ via cross-identity driven manner. After introducing contrastive learning loss (w/ $\mathcal{L}_{CON}$), we achieve a significant improvement in the detail quality of the toonify appearance, as indicated by the arrows (\textit{e.g.} facial wrinkles \textit{etc.}). Natural face$\copyright$\textit{Wojciech Zielonka et al.} (CC BY).}
  \label{fig:L_CON}
\end{figure}

\subsection{Ablation Study}

We focus on the following aspects for ablation studies.

\subsubsection{Relationship between pre-training and fine-tuning} We enumerated the fine-tuned results under the pre-trained models under different epochs (the fine-tuning steps are set $500$ iterations). As shown in Figure~\ref{fig:em}, a larger pre-train epoch corresponds to a more detailed realistic appearance (mean error $2.74$ for epoch $30$), and the fine-tuned stylization appearance also has more details. It can be concluded that fine-tuning will not contribute additional fine-grained gains to the appearance, and an excellent pre-trained model is fundamental for effective stylization.

\subsubsection{Adaptation to prompts} In the previous mention, the format of the used prompt is similar to \textit{"Turn ... into ..."}. To demonstrate robustness to prompts, we show the results by simple prompts, such as \textit{"Kawaii"}, as shown in the Figure~\ref{fig:m-view} (first row). We present each style from multiple view, which are learned from single-view input. Moreover, our method is adaptable to different prompts and can even handle mixed styles like \textit{"Kid + Marble"}. 

\subsubsection{Stylization intensity} The control over stylization intensity is shown in the second row in Figure~\ref{fig:m-view}. We take the ratio of the text control strength and the image guidance strength in the T2I (the larger the ratio, the data provided by T2I is closer to the text description style) as the stylization intensity. The intensity is $0$ representing the facial appearance without stylization. We find that as intensity increases, the appearance is closer to textually descriptive (\textit{e.g.}, the face color becomes whiter) while maintaining the same expressions as the pre-trained one.

\subsubsection{Fine-tuning time} The style trajectory over fine-tuning time are shown in Figure~\ref{fig:m-view} ($3^{rd}$ row). We initialize from the pre-trained model (Time = $0$ min). After more than $2$ minutes, the change in appearance is minor. Generally, the fine-tuning time is set to $5$ minutes.

\subsubsection{Ablation study on $\mathbf{D}_c$} The conditional Tri-plane Gaussian deformation field is used throughout the pre-training and fine-tuning phase. Its main role is to reduce reliance on a large number of MLPs and control facial expressions with conditional inputs. We replace $\mathbf{D}_c$ with the connected MLP layers and repeat the pipeline. As shown in Figure~\ref{fig:D_c}, w/o $\mathbf{D}_c$ is to replace $\mathbf{D}_c$ by MLPs. It can be observed that the introduction of $\mathbf{D}_c$ improves the accuracy of expression for pre-trained realism appearance. This enhancement also positively affects the stylized appearance. Additionally, we present quantitative measurements to verify this perceptual assessment. Specifically, we leverage the average $L_1$ distance on facial key points to evaluate the results corresponding to w/ and w/o $\mathbf{D}_c$ on the pre-trained appearance. Since keypoints cannot be accurately detected on stylized faces, we exclude the fine-tuned appearance from this discussion. Please refer to~\cite{duan2023bakedavatar} for key point distance details. The keypoint distance for w/ and w/o $\mathbf{D}_c$ are \underline{$4.87$} and \underline{$5.90$} respectively, indicating that $\mathbf{D}_c$ leads to more accurate facial movements. 

\begin{figure}[t]
  \centering
  \includegraphics[width=1.\linewidth]{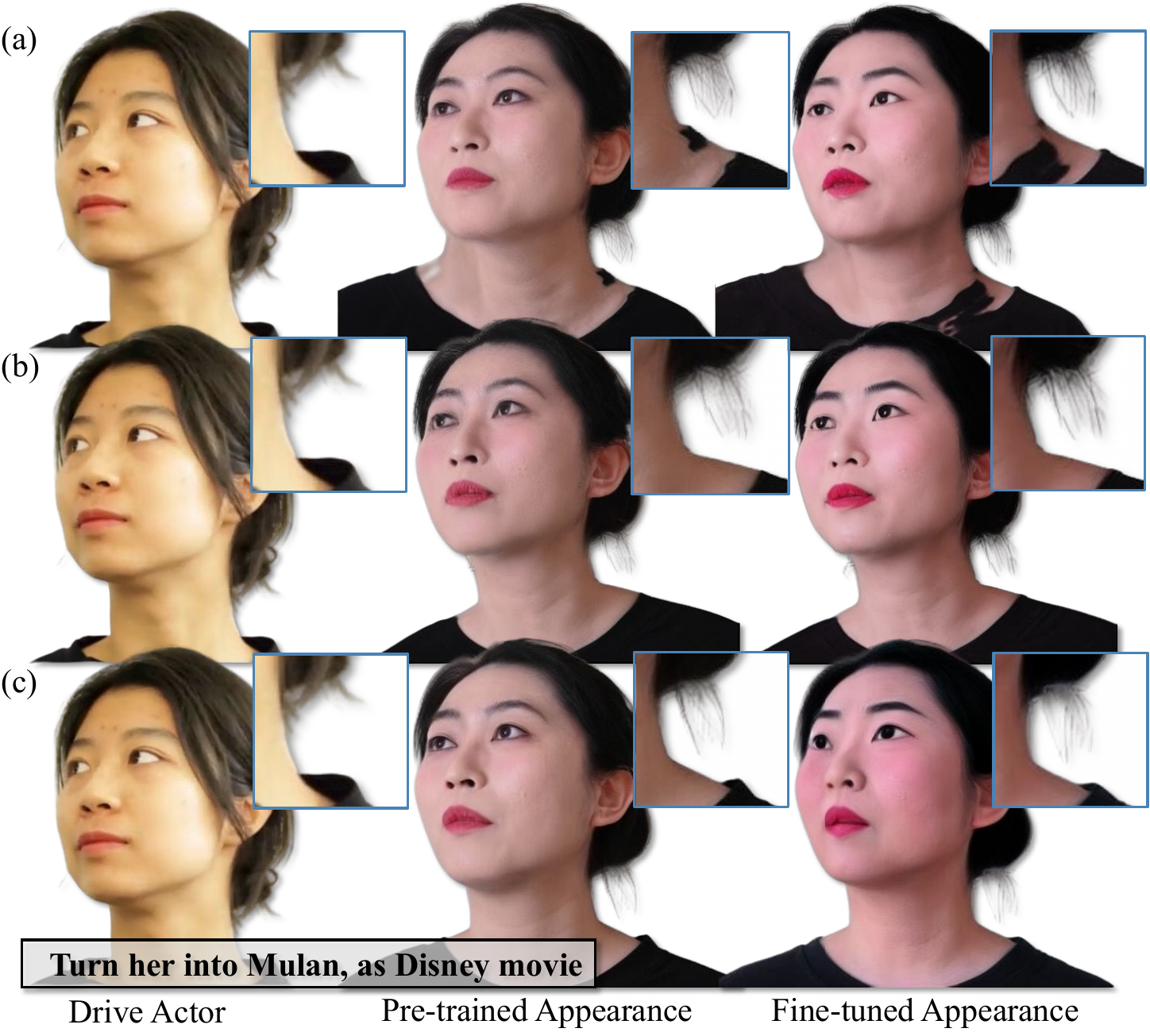}
  \vspace{-0.5cm}
  \caption{The visualize of the ablation study on $w$. (a) is for the hard-parsed motion on head/shoulder, (b) for is the same motion on head/shoulder and (c) for is the adaptively learned $w$. We zoom-in the neck-shoulder transition area to highlight the details. There are artifacts around the neck in (a) due to inconsistent head/shoulder movement, and the shoulder in (b) is displaced synchronously with the head. The influence occurs throughout both the pre-training and fine-tuning phases. Natural face$\copyright$\textit{Yufeng Zheng et al.} (CC BY).}
  \label{fig:w}
\end{figure}

\begin{figure}[t]
  \centering
  \includegraphics[width=1.\linewidth]{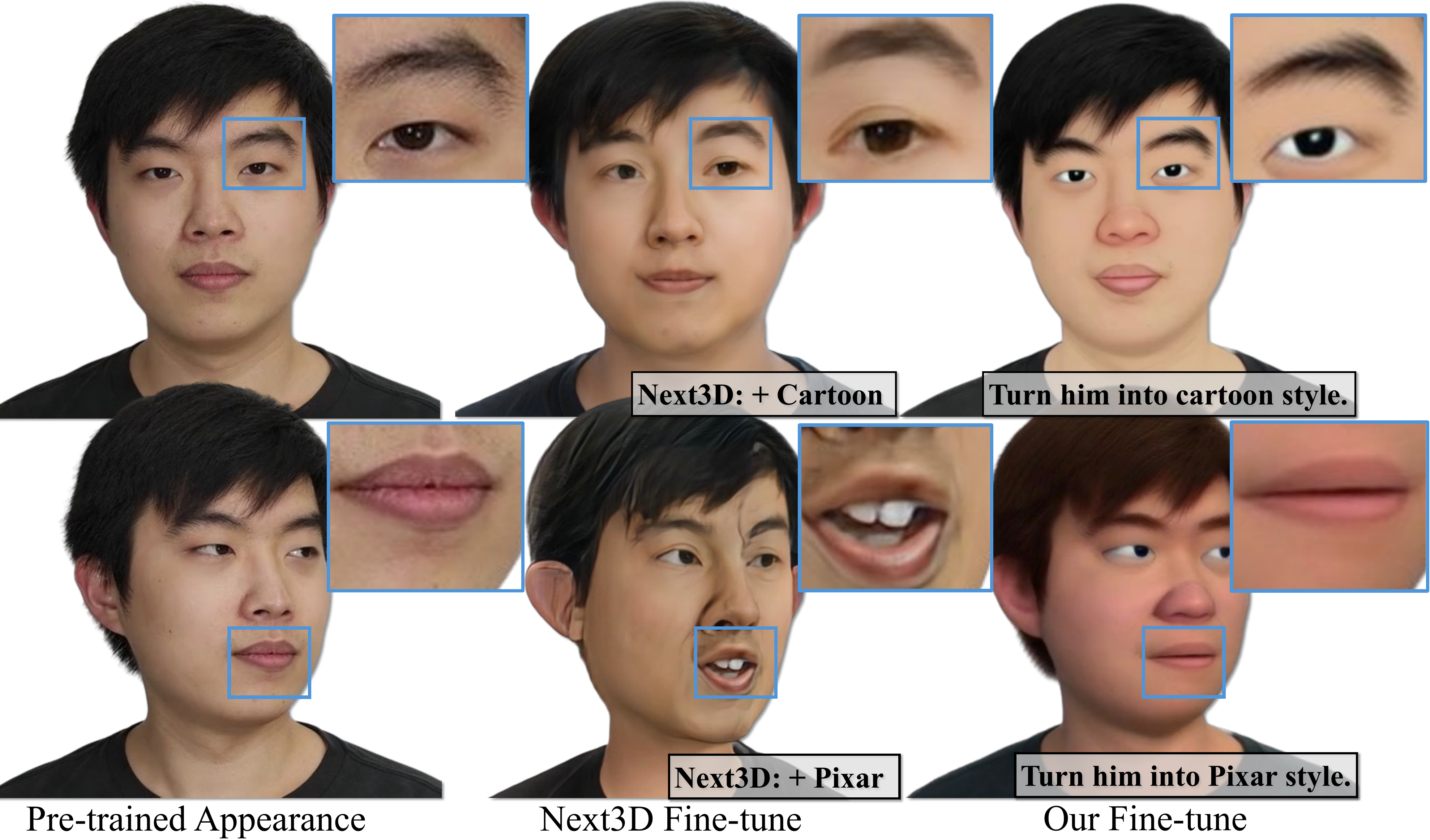}
  \vspace{-0.5cm}
  \caption{The visualize of comparison with Next3D, the PTI (StyleGAN inversion) is used for feature embedding. We find that the results suffer from facial identity drift. Furthermore, the two styles associated with Next3D (Pixar and Cartoon) are not distinct. We apply the prompts ``Cartoon" and ``Pixar" for Next3D, the corresponding prompts for our method are ``Turn him into cartoon style" and ``Turn him into Pixar style". Natural face$\copyright$\textit{Lizhen Wang et al.} (CC BY).}
  \label{fig:oneshot}
\end{figure}

\subsubsection{Ablation study on $\mathcal{L}_{CON}$} The $\mathcal{L}_{CON}$ is used to improve the quality of stylized images through patch-aware contrastive learning during fine-tuning, preventing over-smoothing. We show the fine-tuned appearance with (w/) and without (w/o) it in Figure~\ref{fig:L_CON}. The results without $\mathcal{L}_{CON}$ are smoother and lack the facial details. Additionally, we provide quantitative evaluations: the BRISQUE score with $\mathcal{L}_{CON}$ is $52.5$, while the score is $67.1$ for w/o.

\subsubsection{Ablation study on factor $w$} We introduce the "lazy" factor $w$ in Section~\ref{Photo-Realistic Appearance Pre-training}. The primary assumption behind $w$ is to eliminate the need for separate processing of the shoulder (also referred to as the torso) and the head, as was required in previous works such as AD-NeRF~\cite{guo2021ad}. This adaptive parsing method significantly enhances the robustness of rigid upper-body movement modeling. As illustrated in Figure~\ref{fig:w}, we employ different configurations of $w$ to achieve the transition of rigid movement from canonical space to camera space. (a): A fixed unit rotation matrix $I$ is assigned to $w_2$ in Eq.~\ref{rigid-3} and estimated head rotation matrix for $w_1$, the head and shoulder motion are hard parsed. (b): The head and shoulders share the same estimated rotation matrix. (c): $w$ is adaptively learned. When $w$ is hard-parsed, the head and shoulder movements are uncoordinated, leading to separation as seen in Figure~\ref{fig:w}(a). Conversely, in Figure~\ref{fig:w}(b), the shoulders move in sync with the head, which is not accurate, as the shoulders typically remain still or exhibit only minimal movement. Our method is able to improve stabilize shoulder motion. While shoulder movement can be managed using other techniques, such as Linear Blend Skinning (LBS) around the neck, our approach is specifically designed for the real-time head tracking, as detailed in Section~\ref{Data Acquisition}. It is a trade-off between real-time capable and render quality.

\subsection{Discussion with EG3D Methods}

Some methods related to EG3D \cite{chan2022efficient} also demonstrate their ability in stylization, such as the DeformToon3D~\cite{zhang2023deformtoon3d} and Next3D~\cite{sun2023next3d}. But they requires extra facial alignment and depend on StyleGAN~\cite{karras2019style} inversion for specific identities. And DeformToon3D is built based on StyleSDF~\cite{or2022stylesdf}. The StyleSDF~\cite{or2022stylesdf} does not have the inversion approach so far. In addition, we provide some comparison results with Next3D by PTI~\cite{roich2021pivotal} inversion in Figure~\ref{fig:oneshot}. We find the Next3D suffer from the shift of identity by StyleGAN inversion and facial expression inconsistent. 

\begin{figure}[t]
  \centering
  \includegraphics[width=1.\linewidth]{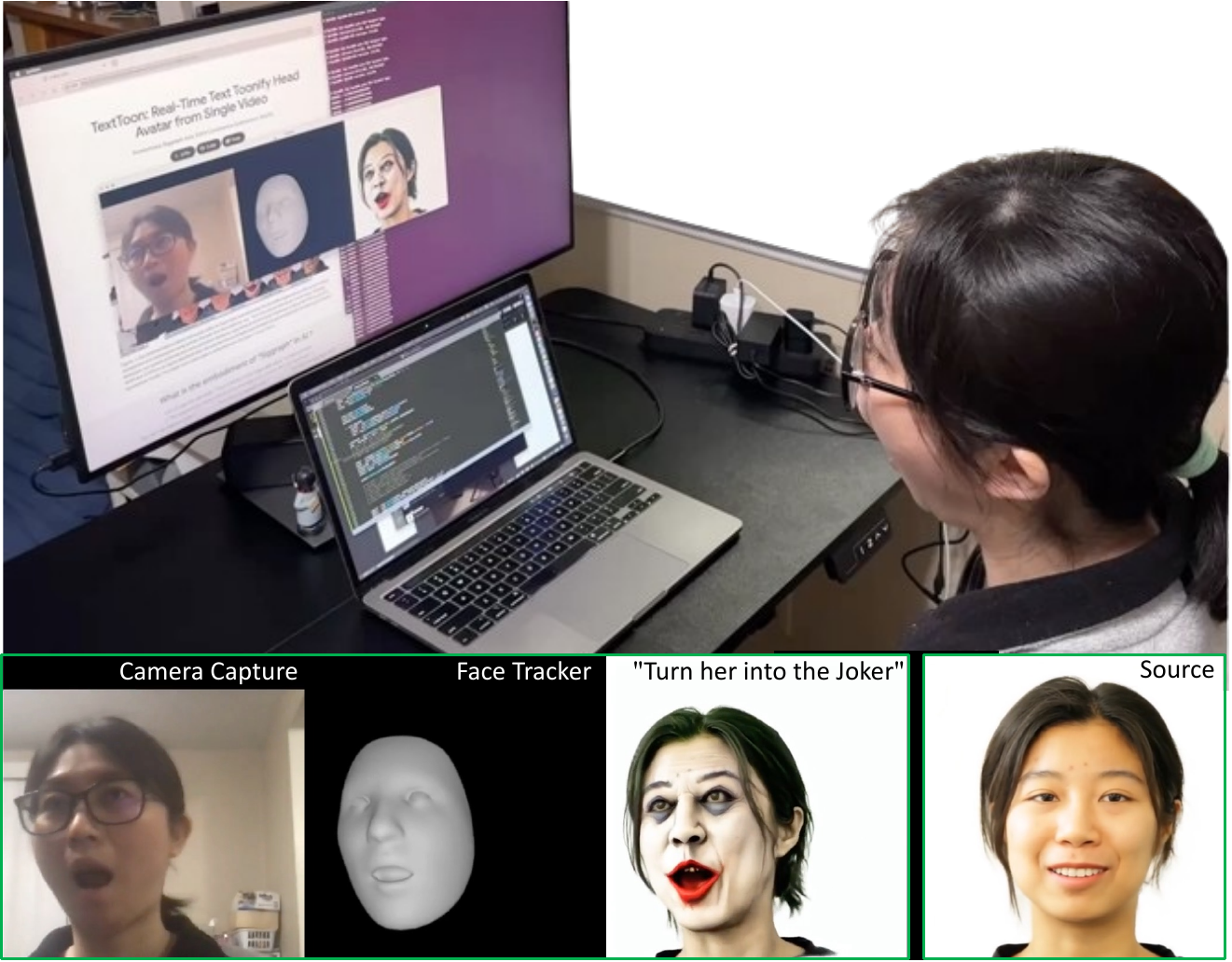}
  \vspace{-0.5cm}
  \caption{The visualize of the real-time application. We use the camera capture appearance to animate the toonified appearance, the prompt ``Turn her into the Joker" is applied for fine-tuning the realistic appearance. The reference source portrait is shown in bottom right corner. Natural face$\copyright$\textit{Yufeng Zheng et al.} (CC BY).}
  \label{fig:realtime}
\end{figure}

\subsection{Real-Time Application}
\label{Real-Time Application}

Benefit from the 3D Gaussian Splatting, the lazy factor $w$ (avoid the separate parsing of head and shoulder) and the CPU optimizations for matrix operations in 3DMM-estimation, we present the real-time application in Figure~\ref{fig:realtime} and the project website. It can achieve the photo-realistic and text-based toonification appearance animation from the camera capture. The system is about $25$ FPS on the daily GPU machine and $18$ FPS on Apple MacBook M1 Chips (with quality compression). Moreover, for the pure inference settings, the speed can be improved to about $48$ FPS. It is worth mentioning that the efficiency of the 3D head tracking algorithm is the bottleneck of our real-time system.

\begin{figure}[t]
  \centering
  \includegraphics[width=1.\linewidth]{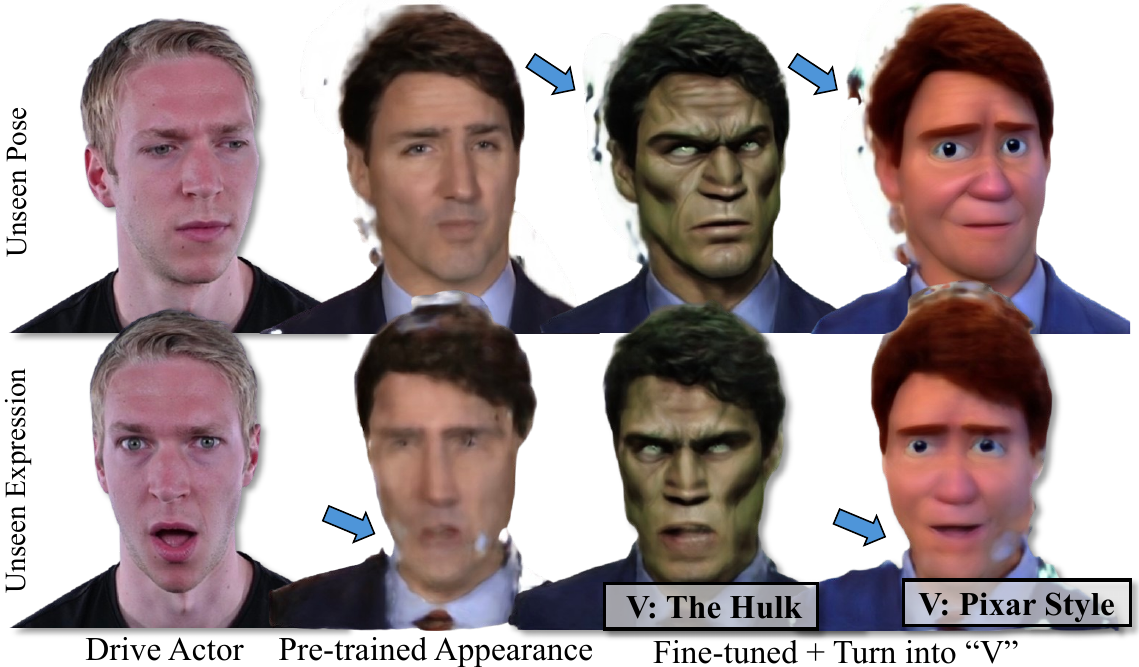}
  \caption{The visualization of the problematic cases. We present the limitations regarding unseen poses and expressions. Artifacts for unseen poses are around the neck and head, while for unseen expressions, blurriness appears near the cheeks. Natural face$\copyright$\textit{Wojciech Zielonka et al.} (CC BY).}
  \label{fig:badcase}
\end{figure}

\section{Discussion and Conclusion}
\label{Discussion and Conclusion}

\subsection{Limitations} There are a few limitations to our methods. First, our stylization relies on the Text2Image module's ability. Achieving fine-grained control such as \textit{"writing `Siggraph' on his/her left cheek"} could be challenging. Second, the pre-training phase is limited by the diversity and quality of the training dataset. Similar to other neural rendering methods (NeRF, 3D Gaussian splatting, and GAN-based methods), our method suffers when generating image frames with out-of-distribution head poses and expressions, as shown in Figure~\ref{fig:badcase} (the drawbacks show in realistic appearance generation and affects toonification, and tend to produce similar artifacts). Third, the input renderings are derived from a 3DMM tracking algorithm, which is not capable of accurately describing detailed expressions, such as precise gaze control or pouting mouth. Finally, our method cannot achieve eye closure in some styles, because the data provided by the Text2Image module does not include eye-closing motions.

\subsection{Technology Abuse Claim} There is a risk of misuse of our method, such as the utilization of racially racism or insulting prompts for the specific portrait synthesis and DeepFakes. We strongly oppose applying our work for such purposes as well as support the development of adversarial generative methods (or called DeepFake detection~\cite{song2022face,song2022adaptive}). And in our paper, we further discuss the drawbacks of stylization synthesis technically, the readers could have a better understanding of this field and know the limitations.

\begin{figure}[t]
  \centering
  \includegraphics[width=1.\linewidth]{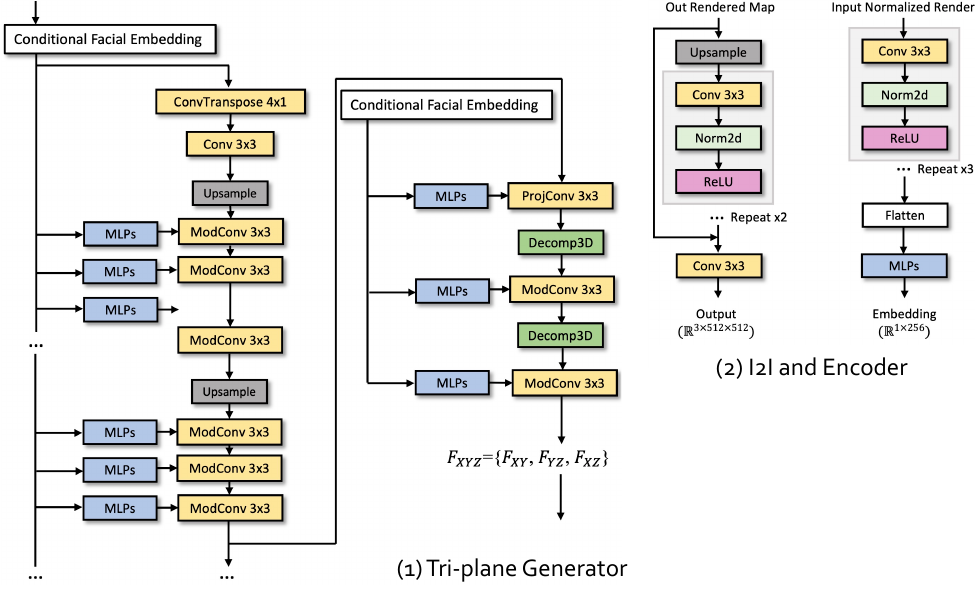}
  \vspace{-0.5cm}
  \caption{The visualize of components details. The conditional facial embedding is a concatenation of encoded orthogonal rendering and the corresponding 3DMM expression coefficients. The size of out rendered map for I2I input is $32 \times 512 \times 512$ and the output size is $3 \times 512 \times 512$. The size of input normalized render is $3 \times 128 \times 128$, and the output of the conditional encoder (embedding) is $1 \times 256$.}
  \label{fig:network_structure}
\end{figure}

\subsection{Technical Details} The details of the components in our pipeline are shown in Figure~\ref{fig:network_structure}. To reduce FLOPs during inference and training, the networks are composed of some light-weight convolution (Conv3$\times$3) and MLP layers. The overall weight of the model is about 49Mb with each components.

\subsection{Conclusion} We present TextToon, a real-time text toonification head avatar generation method from monocular videos. It takes a conditional Tri-plane Gaussian deformation field to learn the Gaussian properties. Meanwhile, we focus on pipeline efficiency and consumer applications, leading to the development of a real-time system, that enables the learned toonify head avatar to be re-animated by other in-the-wild camera captures in real-time. We expect that our method and the principles of real-time system design will inspire further advancements in related approaches.

\bibliographystyle{ACM-Reference-Format}
\bibliography{sample-bibliography}

